\documentclass[conference]{IEEEtran}
\IEEEoverridecommandlockouts
\usepackage{cite}
\usepackage{verbatim}
\usepackage{amsmath,amssymb,amsfonts,bm}
\usepackage{algorithm}
\usepackage{algorithmicx}
\usepackage{enumitem}
\usepackage{amsthm}
\usepackage{algpseudocode}
\usepackage{graphicx}
\usepackage{textcomp}

\usepackage{float}
\usepackage[font=small]{caption}
\usepackage{xcolor}
\usepackage{wrapfig}
\usepackage{hyperref}
\usepackage[font=footnotesize,labelformat=simple]{subcaption}
\usepackage[protrusion=true,expansion=true]{microtype}
\pdfoutput=1

\usepackage{caption}

\def\BibTeX{{\rm B\kern-.05em{\sc i\kern-.025em b}\kern-.08em
    T\kern-.1667em\lower.7ex\hbox{E}\kern-.125emX}}
    
    \makeatletter 
\newcommand\semiHuge{\@setfontsize\semiHuge{22.7}{28.38}}
\makeatother

\begin{document}



\title{Embedding Alignment for Unsupervised Federated Learning via Smart Data Exchange\\
}

\author{\IEEEauthorblockN{Satyavrat Wagle\IEEEauthorrefmark{1}, Seyyedali Hosseinalipour\IEEEauthorrefmark{1}, Naji Khosravan\IEEEauthorrefmark{2}, Mung Chiang\IEEEauthorrefmark{1}, and Christopher G. Brinton\IEEEauthorrefmark{1}}
\IEEEauthorblockA{\IEEEauthorrefmark{1}School of Electrical and Computer Engineering, Purdue University, West Lafayette, IN, USA}
\IEEEauthorblockA{\IEEEauthorrefmark{2}Zillow Group, Seattle, WA, USA}
\IEEEauthorblockA{\IEEEauthorrefmark{1}\{wagles, hosseina, chiang, cgb\}@purdue.edu, \IEEEauthorrefmark{2}najik@zillowgroup.com}\vspace{-6mm}}

\maketitle

\begin{abstract}
Federated learning (FL) has been recognized as one of the most promising solutions for distributed machine learning (ML). In most of the current literature, FL has been studied for \textit{supervised} ML tasks, in which edge devices collect \textit{labeled} data. 
Nevertheless, in many applications, it is impractical to assume existence of labeled data across devices. To this end, we develop a novel methodology, Cooperative Federated unsupervised Contrastive Learning ({\tt CF-CL)}, for FL across edge devices with unlabeled datasets. {\tt CF-CL} employs local device cooperation where data are exchanged among devices through device-to-device (D2D) communications to avoid local model bias resulting from non-independent and identically distributed (non-i.i.d.) local datasets. 
{\tt CF-CL} introduces a \textit{push-pull smart data sharing} mechanism tailored to unsupervised FL settings, in which, each device pushes a subset of its local datapoints to its neighbors as {reserved data points}, and pulls a set of datapoints from its neighbors, sampled through a  probabilistic importance sampling technique. We demonstrate that
{\tt CF-CL} leads to (i) alignment of unsupervised learned latent spaces across devices, (ii) faster global convergence, allowing for less frequent global model aggregations; and (iii) is effective in extreme non-i.i.d. data settings across the devices.

%
\end{abstract}


\vspace{-1mm}
\section{Introduction}\label{intro}
\noindent Many emerging intelligence tasks require training machine learning (ML) models on a distributed dataset across a collection of wireless edge devices (e.g., smartphones, smart cars) \cite{fi11040094}. Federated learning (FL) \cite{9084352,MAL-083} is one of the most promising techniques for this, utilizing the computation resources of edge devices for data processing. Under conventional FL, model training consists of (i) a sequence of local iterations by devices on their individual datasets, and (ii) periodic global aggregations by a main server to generate a global model that is synchronized across devices to begin the next training round.

In this work, we are motivated by two  challenges related to implementation of FL over real-world edge networks. First, device datasets are often non-independent and identically distributed (non-i.i.d.), causing local model bias and degradation in global model performance. Second,  data collected by each device (e.g., images, sensor measurements) is often unlabeled, preventing supervised ML model training. We aim to jointly address these challenges with
a novel methodology for smart data sampling and exchange
across devices \textit{in settings where devices are willing to share their collected data (e.g., wireless sensor measurements, images collected via smart cars)}  \cite{9311906,wang2021device,furl,zhao2018federated,hosseinalipour2022parallel}.



\vspace{-1mm}
\subsection{Related Work and Differentiation}

\subsubsection{FL under non-i.i.d. data}
Researchers have aimed to address the impact of non-i.i.d. device data distributions on FL performance. In~\cite{wang2019adaptive,9562522}, convergence analysis of FL via device gradient diversity-based metrics is conducted, and control algorithms for adapting system parameters are proposed.
In \cite{wang2020optimizing}, a reinforcement learning-based method for device selection is introduced, counteracting local model biases caused by non-i.i.d. data. In~\cite{briggs2020federated}, a clustering-based approach is developed, constructing a hierarchy of local models to capture their diversity. In~\cite{zhao2021federated}, the authors tune model aggregation to reduce local model divergence using a theoretical upper bound.
Works such as \cite{9311906,wang2021device,furl,zhao2018federated,hosseinalipour2022parallel} explored data exchange between devices in FL to improve local data similarities in settings with no strict privacy 
concerns on data sharing.

The emphasis of literature has so far been on supervised ML settings. However, collecting labels across distributed edge devices is impractical for many envisioned FL applications (e.g. images captured by self-driving cars are not generally labeled with object names, weather conditions, etc.) 

\subsubsection{FL for unlabeled data}
A few works have considered unsupervised FL. In \cite{berlo}, a local pretraining methodology was introduced to generate unsupervised device model representations for downstream tasks. The authors in \cite{furl} proposed addressing the inconsistency of local representations through a  dictionary-based method. In \cite{9348203}, the dataset imbalance problem is addressed with aggregation weights at the server defined according to inferred sample densities. In \cite{unsupfl}, unsupervised FL is considered for the case where client data is subdivided into unlabeled sets treated as surrogate labels for training. The authors in  \cite{Li_2021_CVPR} exploited similarities across locally trained model representations to correct the local models' biases.

In this work, we consider \textit{contrastive learning} \cite{1640964,pmlr-v119-chen20j} as our framework for unsupervised ML. Contrastive learning is an ML technique which aims to learn embeddings of unlabeled datapoints that maximize the distance between different points and minimizes it for similar points in the latent space. 

Given the non-i.i.d. data in FL, alignment of locally learned representations is crucial for a faster model convergence. 
Emerging works in supervised FL have shown, when permissible, even a small amount of data exchange among neighboring devices can substantially improve training~\cite{wang2021device}. Subsequently, we exploit device-to-device (D2D) communications as a substrate for improving local model alignment through a novel push-pull data exchange strategy tailored to unsupervised FL.


\subsubsection{Importance sampling}
In the ML community, importance sampling techniques have been introduced to accelerate training through the choice of minibatch data samples \cite{Dong_2018_ECCV}. In FL, by contrast, importance sampling has typically been employed to identify \textit{devices} whose models provide the largest improvement to the global model, e.g., \cite{9413655,wang2021device}. Our work extends the literature of importance sampling to consider inter-device datapoint transfers in a federated setting, where devices exchange their local datapoints through a \textit{probabilistic importance data sampling} to accelerate training speed.

\begin{figure}[t]
    \centering
    \includegraphics[width=.95\columnwidth]{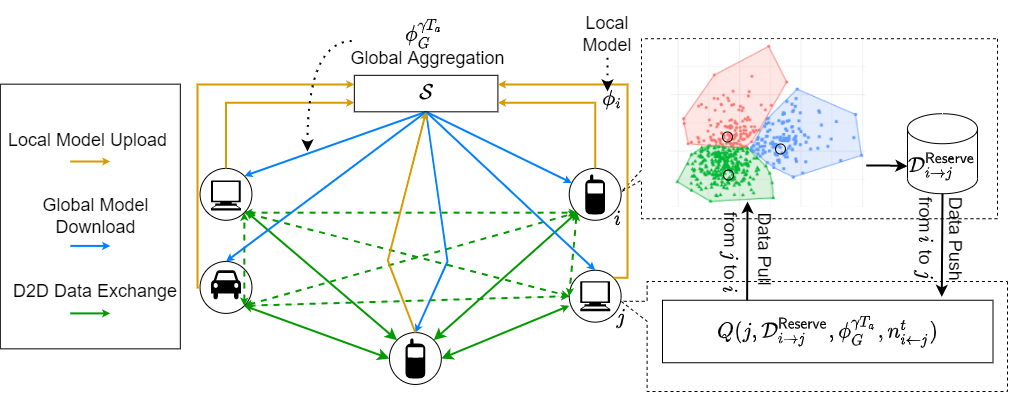}
    \vspace{-1mm}
    \caption{{\tt{CF-CL}} introduces a smart push-pull data transfer mechanism to improve unsupervised FL based on importance data sampling. 
    }
    \label{ss_num_clusters}
    \vspace{-6mm}
\end{figure}

\vspace{-1mm}
\subsection{Summary of Contributions}
Our contributions are threefold: (i) We develop {\tt CF-CL} -- Cooperative Federated unsupervised Contrastive Learning -- a novel method for contrastive FL. {\tt CF-CL} improves training speed via smart D2D data exchanging in settings with no strict privacy 
concerns on data sharing. 
 {\tt CF-CL} provides a general plug-and-play method mountable on current FL methods.
(ii) We introduce a data push-pull strategy based on probabilistic importance data selection in {\tt CF-CL}. We characterize the importance of a remote datapoint via a joint clustering and loss measurement technique to maximize the convergence rate of FL.
(iii) Our numerical results show that {\tt CF-FL} significantly improves FL training convergence compared to baselines. 




\vspace{-.5mm}
\section{System Model and Machine Learning Task}
\noindent An overview of our method is illustrated in Fig~\ref{ss_num_clusters}. In this section, we go over our network model (Sec.~\ref{sec:net}) followed by the ML task for unsupervised FL (Sec.~\ref{sec:ML}).

\vspace{-1.2mm}
\subsection{Network Model of FL with Data Exchange}\label{sec:net}
 We consider a network of a server $S$ and devices/clients gathered via the set $\mathcal{C}$. At each time-step $t$, each device $ i\in \mathcal{C}$ possesses a local ML model parametrized by $\bm{\phi}_i^t\in\mathbb{R}^p$, where $p$ is the number of model parameters. Let $\mathcal{D}_i$ denote the \textit{initial} local dataset at device $ i$. The server aims to obtain a global model $\bm{\phi}_G^{t}$ via aggregating $\{\bm{\phi}_i^t\}_{i\in\mathcal{C}}$, each trained on local dataset and data points received from neighboring devices.

 We represent the communication graph between the devices via $\mathcal{G}=(\mathcal{C},\mathcal{E})$ with vertex set $\mathcal{C}$ and edge set $\mathcal{E}$. The existence of an edge between two nodes $ i$ and $ j$ (i.e., $( i, j)\in \mathcal{E}$) implies a communication link between them.\footnote{This graph can be obtained in practice using the transmit power of the nodes, their distances, and their channel conditions (e.g., see Sec. V of~\cite{9562522}).} We consider an undirected graph where $( i, j)\in \mathcal{E}$ implies $( j, i)\in \mathcal{E}$, $\forall i, j$. We further denote the neighbors of device $ i$ with $\mathcal{N}_i =\{ j: ( i, j)\in \mathcal{E} \}$.  
 
 We consider \textit{cooperation} among the devices~\cite{wang2021device,9311906} in a push-pull setting, where each device $i$ may push a set of local datapoints {\small$\mathcal{D}_{i\rightarrow j}^{\mathsf{Reserve}}$} to another device {\small$j\in \mathcal{N}_i$} (i.e., {\small$\mathcal{D}_{i\rightarrow j}^{\mathsf{Reserve}}\subseteq \mathcal{D}_i$}). {\small$\mathcal{D}_{i\rightarrow j}^{\mathsf{Reserve}}$} is stored at device $j$ as \textit{reserved data points}, which will be later used to determine the important data points to be pulled by device $i$ from device $j$. The set of pulled data points from device $j$ to $i$ ({\small$\mathcal{D}_{i\leftarrow j}^{t}\subseteq \mathcal{D}_j$}) are periodically selected 
based on a probabilistic sampling scheme 
 by a \textit{selection algorithm} {\small$Q(j,\mathcal{D}_{i\rightarrow j}^{\mathsf{Reserve}},\bm{\phi}_G^{t}) = \mathcal{D}_{i\leftarrow j}^t \subseteq \mathcal{D}_j $} detailed in Sec.~\ref{subsec:B}.





\begin{figure}
    \centering
       \hspace{-3.5mm} \includegraphics[width=1.0\columnwidth]{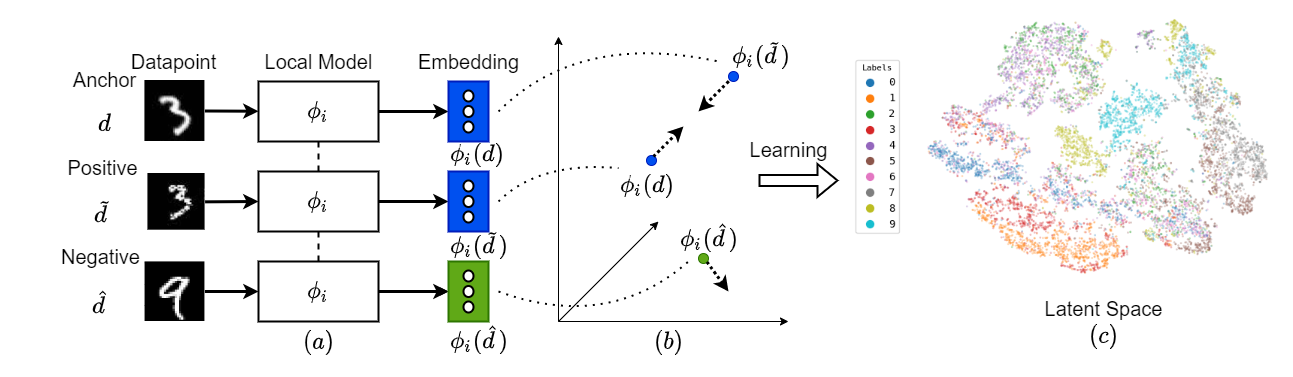}
    \vspace{-2mm}
    \caption{(a) A datapoint (\textit{anchor}), its augmentation (\textit{positive}) and a distinct datapoint (\textit{negative}), are passed through models to obtain embeddings (b). Training maximizes distance between anchor and negative, while minimizing distance between anchor and positive (c).}
    \label{fig:triplet_loss}
    \vspace{-5mm}
\end{figure}

\vspace{-.5mm}
\subsection{Unsupervised  FL Formulation}\label{sec:ML}
We consider an unsupervised learning task, the goal of which is to learn effective \textit{embeddings} of datapoints (i.e., projections of datapoints onto a latent space). To this end, we exploit contrastive learning (CL), which is extensively studied in the centralized ML domain \cite{1640964}, \cite{pmlr-v119-chen20j}. CL obtains embeddings by minimizing the \textit{distance} between similar datapoints while maximizing it between dissimilar datapoints. 
In unsupervised learning, given an anchor datapoint $d$, a similar datapoint (i.e., a positive) is obtained by applying a randomly sampled augmentation function (e.g., image transformations, Gaussian blurs, and color jitter) to the anchor image\cite{pmlr-v119-chen20j}, \cite{moco}. Any distinct datapoint from the anchor is chosen as a dissimilar datapoint (i.e., a negative).
Given model $\bm{\phi}$, margin $m$, anchor $d$, augmented view {\small$\tilde{d} = {F}(d)$}, where ${F}$ is a random augmentation function selected from a set of predefined augmentation functions {\small$\mathcal{F}$} (i.e., {\small$F \in \mathcal{F}$}), and distinct datapoint $\hat{d}$, \textit{triplet loss} $\mathcal{L}$ \cite{Dong_2018_ECCV} is defined for a triplet of datapoints $\{d,\tilde{d},\hat{d}\}$ as 
\vspace{-4mm}

{\small
\begin{equation}
    \label{eqn : triplet_loss}
    \mathcal{L}_{\bm{\phi}}(d,\tilde{d},\hat{d}) \hspace{-.6mm}=\hspace{-.6mm} \max \hspace{-.7mm}\bigg[\hspace{-.1mm}0,\hspace{-.5mm} ||\bm{\phi}(d)-\bm{\phi}(\tilde{d})||_2^2 \hspace{-.9mm}-\hspace{-.9mm} ||\bm{\phi}(d)-\bm{\phi}(\hat{d})||_2^2 \hspace{-.7mm}+\hspace{-.7mm} m\hspace{-.1mm}\bigg]\hspace{-.2mm}.
    \hspace{-.7mm}
\end{equation}
}
\vspace{-4.2mm}

\noindent Triplet loss leads to a latent space in which similar datapoints are closer to one another while dissimilar ones are further away by at least a margin of $m$ as illustrated in
Fig \ref{fig:triplet_loss}.



In our distributed ML setting, we define the goal of unsupervised  FL as identifying a global model $\bm{\phi}_G^\star$ such that 
\vspace{-1.5mm}
\begin{equation}\label{eq:mainProb}
\vspace{-.7mm}
    \bm{\phi}_G^\star = \min_{\bm{\phi}\in\mathbb{R}^p}~ \sum_{d \in \mathcal{D}} \sum_{F \in \mathcal{F}} \sum_{\hat{d} \in \mathcal{D}, \hat{d}\neq d} \mathcal{L}_{\bm{\phi}}(d,\tilde{d},\hat{d}), \tilde{d} = {F}(d),\vspace{-1mm}
\end{equation}
where {\small$\mathcal{D} = \bigcup_{i\in\mathcal{C}} \mathcal{D}_i$} represents the global dataset. The  optimal \textit{global latent space} (e.g., subplot c in Fig.~\ref{fig:triplet_loss}) is the one in which the positive and anchors are closer to each other and further away from negative samples across the \textit{global dataset}. 

To achieve faster convergence to such a latent space in a federated setting, where data distribution across devices is non-i.i.d., \textit{alignment} between \textit{local} latent spaces during training is crucial. We propose to speed up this cross device alignment by smart data transfers. Intuitively, when the data across the devices is homogeneous (i.e., i.i.d.), given a unified set of local models, the local latent spaces are aligned, under which the model training across the devices would exhibit a fast convergence. Thus, given the non-i.i.d. data across the devices, we select and share a set of \textit{important} datapoints across the devices, which result in the best alignment of their local models.

\vspace{-1.2mm}
\section{Cooperative Federated
\label{sec:coop_fed}
Unsupervised Contrastive Learning ({\tt CF-CL})}
\noindent In this section, we first introduce the ML model training process of {\tt CF-CL} in Sec.~\ref{sec:train}. We then detail its efficient cooperative data transfer across the nodes in Sec.~\ref{subsec:B}.

\vspace{-1.2mm}
\subsection{Local/Global Model Training in {\tt CF-CL}}\label{sec:train}
\vspace{-.3mm}
In {\tt CF-CL},~\eqref{eq:mainProb} is solved through a sequence of global model aggregations indexed by {\small$\gamma \in \mathbb{Z}^+$} such that local models (see Sec.~\ref{sec:net}), are aggregated at time-steps {\small$t \in \{\gamma T_a\}_{\gamma \in \mathbb{Z}^+}$}, where {\small$T_a$} is the aggregation interval. The system is trained for {\small$T$} time-steps, where in each time-step, each device {\small$i\in\mathcal{C}$} conducts one mini-batch stochastic gradient descent (SGD) iteration


The data exchange process is a combination of a single initial push of data from each device {\small$i\in\mathcal{C}$} to its neighbors (forming {\small$\mathcal{D}_{i\rightarrow j}^{\mathsf{Reserve}}$},
{\small$j\in\mathcal{N}_i$}).\footnote{The pushed data will be used for the purposes of importance calculation.}
This is followed by a \textit{periodic} pull of data indexed by {\small$\tau \in \mathbb{Z}^+$}, at time-steps {\small$t \in \{\tau T_p\}_{\tau \in \mathbb{Z}^+}$}, where {\small$T_p$} is the data pull {interval}. At each {\small$t=\tau T_p$}, {\small$\tau\in\mathbb{Z}^+$},
each device $i$ requests pulling {\small$n^t_{i\leftarrow j}$} datapoints, {\small$\mathcal{D}^{t}_{i\leftarrow j}$}, from device {\small$j\in\mathcal{N}_i$}. In practice, the number of pulled datapoints across devices (i.e., data exchange budget) {\small$\{n^t_{i\leftarrow j}\}_{i\in\mathcal{C},j\in\mathcal{N}_i}$} can be determined according to bandwidth and channel state condition (CSI) across the network devices. The focus of this work is not designing {\small$\{n^t_{i\leftarrow j}\}_{i\in\mathcal{C},j\in\mathcal{N}_i}$}, rather we assume known values for data exchange budgets and focus on smart data sampling.

We assume that each device has a buffer of limited size to store remote data, and hence purges any remote datapoint pulled in the previous iterations {\small$\tau'$}, where {\small$\tau' T_p < \tau T_p$}, before pulling data at {\small$\tau T_p$}.\footnote{Our method readily applies to a setting in which devices have unlimited buffer sizes and accumulate the pulled data points.}. The push-pull procedure is detailed in~Sec.~\ref{subsec:B}.



At each time-step {\small$t$} after the last data pull {\small$\tau$} (i.e., {\small$ \tau T_p\leq t < (\tau+1) T_p$}), given the data points stored at each device {\small$\mathcal{D}_i^{t}=\mathcal{D}_i^{\tau T_p}\triangleq \mathcal{D}_i \cup \tilde{\mathcal{D}}_i^{\tau T_p}$} (i.e., its initial data points {\small${\mathcal{D}_i}$} and the data points pulled from its neighboring devices {\small$\tilde{\mathcal{D}}^{\tau T_p}_i=\cup_{j\in\mathcal{N}_i} \mathcal{D}^{\tau T_p}_{i\leftarrow j}=\bigcup_{j\in\mathcal{N}_i} Q(j,\mathcal{D}_{i\rightarrow j}^{\mathsf{Reserve}},\bm{\phi}_G^{\gamma T_a})$}, where {\small$\bm{\phi}_G^{\gamma T_a}$} is the most recent global model at time $t$, i.e., {\small$\gamma T_a < t < (\gamma + 1)T_a$}), we conduct local model training at device $i$ to  obtain a local model that minimizes the local triplet loss function as follows:
\vspace{-3mm}
\begin{equation}
\label{eqn:local_loss}
   \bm{\phi}^\star_i = \min_{\bm{\phi}} \mathcal{L}_{\bm{\phi}}(d,\tilde{d},\hat{d}), ~ d \in \mathcal{D}_i^{t}, \tilde{d} = {F}(d), \hat{d} \in \mathcal{D}_i^{t}.
   \vspace{-1mm}
\end{equation}



To solve~\eqref{eqn:local_loss}, devices undergo local model updates via SGD iterations.
At each time-step $t$, given local model $\bm{\phi}_i^{t}$ and a mini-batch of triplets {\small$\mathcal{B}_i^{t}  = \big\{(d,\tilde{d},\hat{d}): {d} \in \mathcal{D}^{{\tau T_p}}_i, \tilde{d} =  {F}(d), {F} \in \mathcal{F} , \hat{d}\in \mathcal{D}^{{\tau T_p}}_i\big\}$}, where {\small$\tau$} is the index of the last data pull (i.e., {\small$\tau T_p < t < (\tau+1)T_p$}), device $i$ updates its local model as
\vspace{-.8mm}
\begin{equation}
    \label{eq:local_update}
    \bm{\phi}_i^{t+1} = \bm{\phi}_i^{t} -\alpha  \sum_{(d,\tilde{d},\hat{d}) \in \mathcal{B}_i^t}{\nabla_{\bm{\phi}_i^{t}}\mathcal{L}_{{\bm{\phi}_i^{t}}}(d,\tilde{d},\hat{d})}\big/{|\mathcal{B}_i^{t}|},
    \vspace{-.8mm}
\end{equation}
where $\alpha$ is the learning rate.

To solve~\eqref{eq:mainProb}, using the local model obtained via~\eqref{eq:local_update} after every $T_a$ local model training rounds, the
local models of the devices are aggregated (at $t\in \{\gamma T_a\}_{\gamma\in\mathbb{Z}^+}$) at server to generate a global model $\bm{\phi}^{{t}}_G$. The server aggregates the local models in proportion to the average cardinality of local datapoints since the last aggregation round $(\gamma-1)$, $|\mathcal{D}_i^{(\gamma-1:\gamma)}|\triangleq \sum_{t \in [(\gamma-1) T_a+1,\gamma T_a]}{|\mathcal{D}_i^t|}/{T_a}$, $\forall i$, as follows:
\begin{equation}
\vspace{-.8mm}
    \label{eq:aggregation}
   \hspace{-3mm} \bm{\phi}^{t}_G = \frac{1}{\sum\limits_{i\in\mathcal{C}}|\mathcal{D}_i^{(\gamma-1:\gamma)}|}\sum_{i\in\mathcal{C}} \bm{\phi}^{t}_i{|\mathcal{D}_i^{(\gamma-1:\gamma)}|},~t = \gamma T_a,\hspace{-.7mm}~\gamma\in\mathbb{Z}^+  \hspace{-.7mm}.  \hspace{-3mm}
    \vspace{-.8mm}
\end{equation}
 Global model $\bm{\phi}_G^t$ is then broadcast across all devices and used to synchronize/override local models $\bm{\phi}_i^{t}\leftarrow\bm{\phi}^{t}_G,~t = \gamma T_a,~\gamma\in\mathbb{Z}^+,~ \forall i\in\mathcal{C}$, and is used for subsequent local training as  in~\eqref{eq:local_update}.
 
 
  
  The pseudo-code of {\tt CF-CL} is given in
Algorithm~\ref{alg:flde}, summarizing data push (line \ref{state:2}-\ref{state:4}), data pull (line \ref{state:8}-\ref{state:update_dp}), 
local training (line \ref{state:update_local_model}), and model aggregation (line \ref{state:update_global_model}) processes.
We next detail our push and pull data exchange processes.
\vspace{-1mm}


 \begin{algorithm}[h]
    \caption{ {\tt CF-CL} Procedure at each Device $i\in\mathcal{C}$}
    \label{alg:flde}
       {\small
\begin{algorithmic}[1]
    \State \textbf{Input:} $i, T_a,T_p, \alpha,$ $\mathcal{D}_i$, $\{n^t_{i\leftarrow j}\}_{j\in\mathcal{N}_i}$
    \State Device $i$ receives the initial global model from the server  $\bm{\phi}_G^{0}$
    \State Device $i$ performs K-means clustering with $K=K^{\mathsf{Reserve}}_{i \rightarrow j}$ on $\mathcal{D}_i$ \label{state:2}
    \State Device $i$ samples reserve data $\mathcal{D}^{\mathsf{Reserve}}_{i \rightarrow j}$ by choosing $K^{\mathsf{Reserve}}_{i \rightarrow j}$ datapoints closest to centroids.
    \State Device $i$ pushes reserve data $\mathcal{D}^{\mathsf{Reserve}}_{i \rightarrow j}$ to each of its neighboring devices $j, ~ j \in \mathcal{N}_i$\label{state:4}
    \For{$t=1 $ \textrm{to} $ T$}
        \If{$t = \tau T_p, \tau  \in \mathbb{Z}^+$}
            \For{$ j \in \mathcal{N}_i$}
                \State  Device $i$ requests pulling $n_{i \leftarrow j}^t$ datapoints from $j$ \label{state:8}
                \State  Device $j$  transmits datapoints to device $i$ by invoking \label{state:update_dp} $\textproc{Q}(j,\mathcal{D}^{\mathsf{Reserve}}_{i \rightarrow j},\bm{\phi}_G^{t},n_{i \leftarrow j}^t)$ (see Algorithm~\ref{algo:sampling})
            \EndFor
        \EndIf
        \\
         Update local model $\bm{\phi}_i^t$ according to (\ref{eq:local_update}) \label{state:update_local_model}
        \If{$t = \gamma T_a, \gamma \in \mathbb{Z}^+$}
            \State Send local model to the server, which updates global model $\bm{\phi}_G^t$ according to (\ref{eq:aggregation}) \label{state:update_global_model}
        \EndIf
    \EndFor
\end{algorithmic}
}
\end{algorithm}



\vspace{-4mm}
\subsection{Representation Alignment via Smart Data Push-Pull}\label{subsec:B}


 In FL, devices' datasets are non-i.i.d., thus local models get biased to local data distributions. We propose a smart push-pull data schema for better alignment of local embedding spaces. 
\subsubsection{Smart Data Push}\label{sec:PushKmeans}
The data exchange gets kicked off by each device $i$ pushing a set of \textit{representative} datapoints to each of its neighbors $j$, stored as reserve datapoints {\small$\mathcal{D}_{i\rightarrow j}^{\mathsf{Reserve}}$} at $j$. Ideally, these datapoints should best capture the \textit{modes} of the local data distribution. Reserve datapoints will later be used to identify important datapoints that contribute the most to cross-device embedding alignment.
Letting {\small$K^{\mathsf{Reserve}}_{i \rightarrow j}\triangleq|\mathcal{D}_{i\rightarrow j}^{\mathsf{Reserve}}|$}, the set of reserved datapoints is calculated as
\vspace{-1mm}
    \begin{align}
        \label{eq:approx_reserve}
        \mathcal{D}^{\mathsf{Reserve}}_{i \rightarrow j} = \{d: d \sim \mathcal{D}_i\},~ |\mathcal{D}^{\mathsf{Reserve}}_{i \rightarrow j}| = K^{\mathsf{Reserve}}_{i \rightarrow j},~j\in\mathcal{N}_i.
    \end{align}

We use {{\small$K$}-means++} \cite{kmeans} clustering, with {\small$K=K^{\mathsf{Reserve}}_{i \rightarrow j}$} at each device $i$ and include the centers of clusters/centroids in
{\small$\mathcal{D}^{\mathsf{Reserve}}_{i \rightarrow j}$}, {\small$j\in\mathcal{N}_i$}.
This will have a dramatic performance increase compared to random data sampling, especially upon having a small size of reserved datapoints (Sec.~\ref{experiments}).



\subsubsection{Smart Data Pull}

We next aim to develop $Q$, determining datapoints pulled by each device {\small$i\in\mathcal{C}$} from device {\small$ j\in\mathcal{N}_i$}. To improve the efficiency and make our methodology practical upon having large local dataset sizes, at each global aggregation time {\small$t=\gamma T_a$}, {\small$\forall \gamma$}, we first approximate local dataset of transmitter device $j$ by uniform sampling a fixed number {\small$K^{\mathsf{Approx}}_j$} of local datapoints
constituting the set {\small$\mathcal{D}^{ t,\mathsf{Approx}}_j$} as
    \vspace{-5.5mm}
   
   { \begin{align}
        \label{eq:approx_local}
       \hspace{-2mm} \mathcal{D}^{ t, \mathsf{Approx}}_j = \{d: \hspace{-.5mm} d \sim \mathcal{D}^{t}_j\},\hspace{-.5mm}~|\mathcal{D}^{ t,\mathsf{Approx}}_j| \hspace{-.5mm}=\hspace{-.5mm} K^{\mathsf{Approx}}_j,\hspace{-.5mm}~t=\gamma T_a. \hspace{-2mm}
    \end{align}
    }
    \vspace{-5.5mm}
    
{\small$\mathcal{D}^{t,\mathsf{Approx}}_j$} constitutes the set of candidate datapoints at device {\small$j$} for transmission to neighboring devices.

At each data pull instance {\small$\tau$}, occurring between two global aggregation rounds {\small$\gamma$} and {\small$\gamma +1$} (i.e., {\small$\gamma T_a \leq \tau T_p < (\gamma+1) T_a $}), the data pulled by device {\small$i$} from device {\small$j$} is obtained by execution of $Q$ and denoted by
 {\small$ \mathcal{D}^{\tau T_p}_{i\leftarrow j}= Q(j,\mathcal{D}_{i\rightarrow j }^{\mathsf{Reserve}},\bm{\phi}_G^{\gamma T_a})\subseteq \mathcal{D}^{ t, \mathsf{Approx}}_j$}.
Design of $Q$ ideally leads to the faster convergence of global models {\small$\bm{\phi}_G^{t}$} to {\small$\bm{\phi}^{\star}_G$} by sampling and pulling datapoints that are \textit{important} (i.e., those that accelerate the convergence of local models while avoiding local model bias).
Global model {\small$\bm{\phi}_G^{\gamma T_a}$}  is used in $Q$ to determine the most effective datapoints from device {\small$j$} to minimize device {\small$i$}'s bias to its local dataset. This will lead to alignment of representations generated across the devices, accelerating the global model convergence.

 To perform data pull between each pair of devices {\small$(i,j)$}, we propose a \textit{two-stage probabilistic importance sampling} procedure, consisting  of a macro and a micro sampling steps.
 In \textit{macro sampling}, we  obtain the embeddings of all datapoints in {\small$\mathcal{D}_{i\rightarrow j}^{\mathsf{Reserve}}$ and $\mathcal{D}^{t, \mathsf{Approx}}_j$} using {\small$\bm{\phi}_G^{t}$}, and perform K-means++ to obtain clusters of embeddings {\small$\mathcal{L}^{t}_{i\leftarrow j}$}. Then, we assign a sampling probability to each of the $K$-means clusters (i.e., cluster-level importance). In particular,
 at device {\small$j$}, we obtain the macro probability of sampling of cluster {\small$\ell \in \mathcal{L}^{t}_j$} as
 \vspace{-1mm}
 \begin{equation}\label{eq:ProbBasic}
     P^{t,\mathsf{Macro}}_{i\leftarrow j} ({\ell}) =  \frac{X^{t,\mathsf{Macro}}_{i\leftarrow j} ({\ell})}{ \sum_{\ell\in \mathcal{L}^{t}_{i\leftarrow j}} X^{t,\mathsf{Macro}}_{i\leftarrow j} ({\ell})},~t=\tau T_p,
      \vspace{-1mm}
 \end{equation}
 where
  \vspace{-2.5mm}
 \begin{equation}\label{eq:xX}
   X^{t,\mathsf{Macro}}_{i\leftarrow j} ({\ell}) \triangleq \frac{K^{t,\mathsf{Approx}}_{i\leftarrow j}(\ell)}{{K^{t,\mathsf{Approx}}_{ j}(\ell)+K^{t,\mathsf{Push}}_{i\rightarrow j}(\ell)}}.
    \vspace{-1mm}
 \end{equation}
 In~\eqref{eq:xX}, {\small$K^{t,\mathsf{Approx}}_{i\leftarrow j}(\ell)$} is the number samples of {\small$\mathcal{D}^{t, \mathsf{Approx}}_j$} located in cluster {\small$\ell$} (i.e., {\small $\sum_{\ell\in \mathcal{L}^{t}_j}K^{t,\mathsf{Approx}}_{i\leftarrow j}(\ell)=K^{\mathsf{Approx}}_j,\forall t$}) and $K^{t,\mathsf{Push}}_{i\rightarrow j}(\ell)$ is the number samples of $\mathcal{D}^{t, \mathsf{Reserve}}_{i\rightarrow j}$ located in cluster $\ell$ (i.e., {\small$\sum_{\ell\in \mathcal{L}^{t}_j}K^{t,\mathsf{Push}}_{i\rightarrow j}(\ell) = K^{\mathsf{Reserve}}_{i\rightarrow j},\forall t$}).
Intuitively, $P^{t,\mathsf{Macro}}_{i\leftarrow j} ({\ell}) $ in \eqref{eq:ProbBasic} results in sampling larger number of datapoints from clusters containing a higher ratio of datapoints in the transmitter $j$ to reserved datapoints of receiver $i$  (i.e., clusters with less similar datapoints to existing ones in the receiver), and thus promotes homogeneity of datasets upon data transfer.


  In \textit{micro sampling}, at  {\small$t=\tau T_p$}, once sampling probabilities of clusters are calculated via \eqref{eq:ProbBasic}, we obtain  sampling probabilities of individual datapoints (i.e., data-level importance).  We assign a probability {\small$P^{t,\mathsf{Micro}}_{i\leftarrow j}  ({\hat{d}})$} to data point {\small$\hat{d}$} in cluster {\small$\ell$} (i.e., {\small$\hat{d}\in \ell$, $\ell\in\mathcal{L}^{t}_{i\leftarrow j}$}) according to the average/expected loss 
  when it is used as a negative with datapoints in $\mathcal{D}_{i\rightarrow j}^{\mathsf{Reserve}}$ used as anchors
  
\vspace{-3.5mm}

\vspace{-1mm}

{\small 
\begin{equation}
\vspace{-2mm}
    \label{eq:exp_loss}
   \hspace{-2mm}\mathbb{E}_{\hspace{-.2mm}d \sim \mathcal{D}^{\mathsf{Reserve}}_{i\rightarrow  j}} \hspace{-.2mm}\hspace{-1mm} \left[\hspace{-.2mm}\hspace{-.3mm}\mathcal{L}_{\hspace{-.3mm}\bm{\phi}^{\gamma T_a}_G}(\hspace{-.2mm}d,\hspace{-.2mm}\tilde{d},\hspace{-.2mm}\hat{d})\hspace{-.3mm}\hspace{-.2mm}\hspace{-.1mm}\right] \hspace{-.2mm}\hspace{-1mm}=\hspace{-1mm} \frac{\hspace{-.6mm}\sum_{d \in \mathcal{D}^{\mathsf{Reserve}}_{i\rightarrow  j}}\hspace{-.2mm} \mathcal{L}_{\hspace{-.2mm}\bm{\phi}^{\gamma T_a}_G}\hspace{-.5mm}(\hspace{-.2mm}d,\hspace{-.2mm}\tilde{d},\hspace{-.2mm}\hat{d})}{K^{\mathsf{Reserve}}_{i\rightarrow  j}}\hspace{-.2mm},\hspace{-.3mm} \tilde{d} \hspace{-.2mm}\hspace{-.2mm}=\hspace{-.2mm}\hspace{-.2mm} F(d),\hspace{-2mm}
\end{equation}
}
\vspace{-3.5mm}

\noindent 
and compute the probability of selection of datapoint $\hat{d}\in\ell$ as 
\vspace{-4mm}

{\small 
\begin{equation}
    \label{eqn:imp_sampling}
  \hspace{-3mm} P^{t,\mathsf{Micro}}_{i\leftarrow j}  ({\hat{d}}) = 
    \frac{\exp\left(\lambda^t \cdot \mathbb{E}_{d \sim \mathcal{D}^{\mathsf{Reserve}}_{i\rightarrow  j}}\hspace{-1mm}\left[\mathcal{L}_{\bm{\phi}^{\gamma T_a}_G}(d,\tilde{d},\hat{d})\right]\right)}{  \hspace{-2mm} \sum_{\hat{d}' \in  \ell} \exp\left(\lambda^t \cdot \mathbb{E}_{d \sim \mathcal{D}^{\mathsf{Reserve}}_{i\rightarrow  j}} \hspace{-1mm} \left[\mathcal{L}_{\bm{\phi}^{\gamma T_a}_G}(d,\tilde{d},\hat{d}')\right]\right)}.  \hspace{-3mm} 
\end{equation}}
\vspace{-2.5mm}

\noindent In~\eqref{eqn:imp_sampling}, {\small$\lambda^t$} is the \textit{selection temperature}, tuning the selection probability of samples with different loss values. 
We introduced {\small$\lambda^t$}  to make our selection algorithm robust against the homogeneity of loss, which occurs during the later stages of training.  Considering~\eqref{eqn:imp_sampling}, our selection algorithm improves the model training performance by prioritizing local datapoints to transmit which produce a higher loss at the receiver (measured via their loss over {\small$\mathcal{D}_{i\rightarrow j }^{\mathsf{Reserve}}$}). Finally, the probability of sampling of each datapoint {\small$\hat{d}$} belonging to arbitrary cluster {\small$\ell$} is given by
\vspace{-1mm}
\begin{equation}\label{final}
    P^{t}_{i\leftarrow j}  ({\hat{d}})= P^{t,\mathsf{Micro}}_{i\leftarrow j}  ({\hat{d}})\times P^{t,\mathsf{Macro}}_{i\leftarrow j} ({\ell}), ~\hat{d}\in {\ell}.
    \vspace{-4mm}
\end{equation}
\vspace{-3mm}

\begin{algorithm}[h] 
    \caption{Data Pull by Device $i$ from Device   $j\in\mathcal{N}_i$}
    \label{algo:sampling}
    {\small
\begin{algorithmic}[1]
     \Function{Q}{$j,\mathcal{D}_{i \rightarrow j}^{\mathsf{Reserve}},\bm{\phi}_G^{\gamma T_a},n^t_{i\leftarrow j}$}\\
     ~~~\label{smp_2} Transmitter $j$ approximates local dataset  $\mathcal{D}_j^{t,\mathsf{Approx}}$ as (\ref{eq:approx_local})\label{state:approx_local}
    \Statex \underline{\textit{Macro Importance:}} \\
    ~~~\label{macro} Transmitter $j$ performs Kmeans++ on $\mathcal{D}_{i\leftarrow j}^{\mathsf{Reserve}}$ and $\mathcal{D}^{t, \mathsf{Approx}}_j$\\
    ~~~~\label{macro2} Transmitter $j$ obtains sampling probability $[ P^{t,\mathsf{Macro}}_{i\leftarrow j} ({\ell})]_{\ell\in\mathcal{L}^{t}_{i\leftarrow j}}$ via \eqref{eq:ProbBasic}
    \Statex  \underline{\textit{Micro Importance:}} \\
    ~~~\label{micro}Transmitter $j$ calculates importance of each datapoint $\hat{d}$ in cluster $\ell \in\mathcal{L}^{t}_{i\leftarrow j}$ by \eqref{eqn:imp_sampling}
    \Statex \underline{\textit{Sampling}:}\\
    ~~~\label{smp} Transmitter $j$ samples ${n^t_{i\leftarrow j}}$ datapoints according to probabilities obtained in \eqref{final} and transmits them to device $i$
    \EndFunction
\end{algorithmic}
}
\end{algorithm}

\vspace{-4mm}
Our selection strategy $Q$ is summarized in Algorithm~\ref{algo:sampling} (incorporated into {\tt{CF-CL}} in Algorithm \ref{alg:flde}), where at each data pull instance $\tau$ we first estimate the distribution of local dataset of transmitter (line \ref{smp_2}). We then calculate the macro (lines \ref{macro}-\ref{macro2}) and micro importances (line \ref{micro}), and finally sample ${n^t_{i\leftarrow j}}$ datapoints for transmission (line~\ref{smp}).




\begin{figure*}	
	\centering
	\begin{subfigure}[t]{0.32\textwidth}
		\centering
		\includegraphics[width=\linewidth,height=3cm]{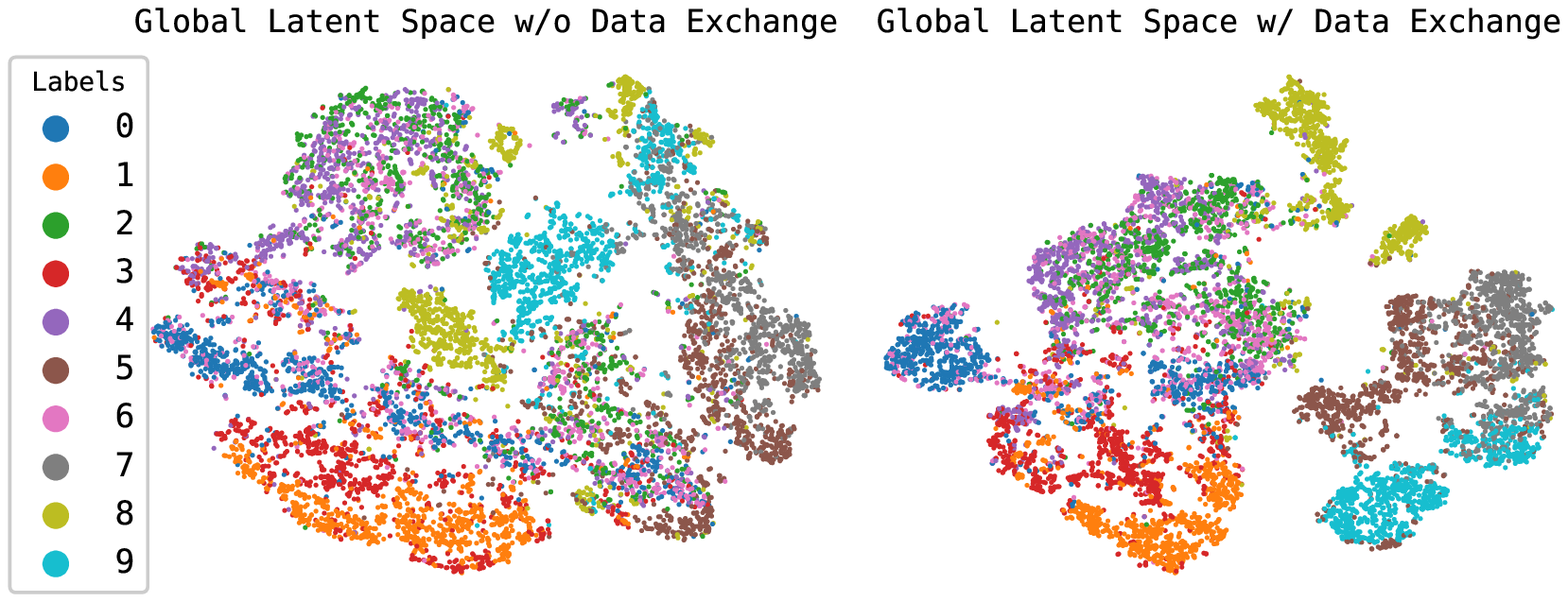}
		\caption{}\label{fig:3a}		
	\end{subfigure}
	\begin{subfigure}[t]{0.31\textwidth}
		\centering
		\includegraphics[width=\linewidth,height=3cm]{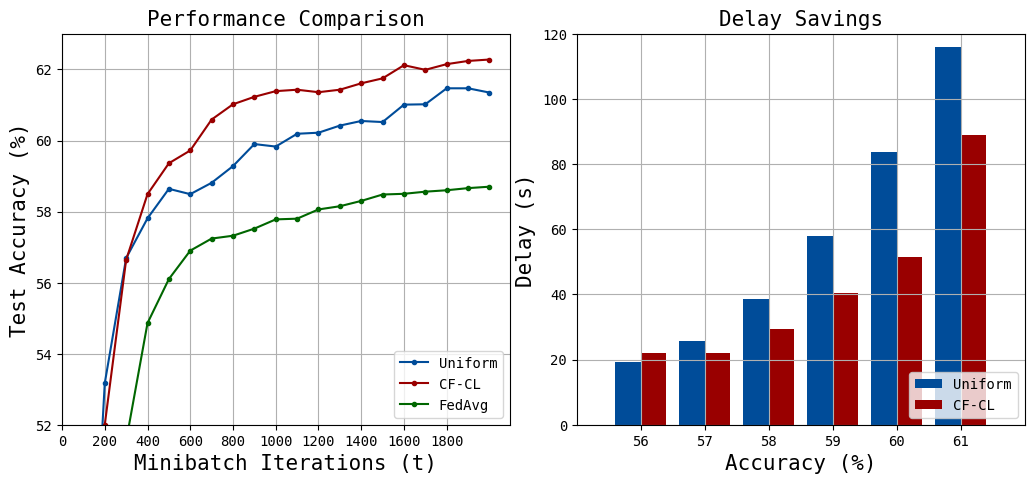}
		\caption{}\label{fig:3b}
	\end{subfigure}
	\begin{subfigure}[t]{0.31\textwidth}
		\centering
		\includegraphics[width=\linewidth,height=3cm]{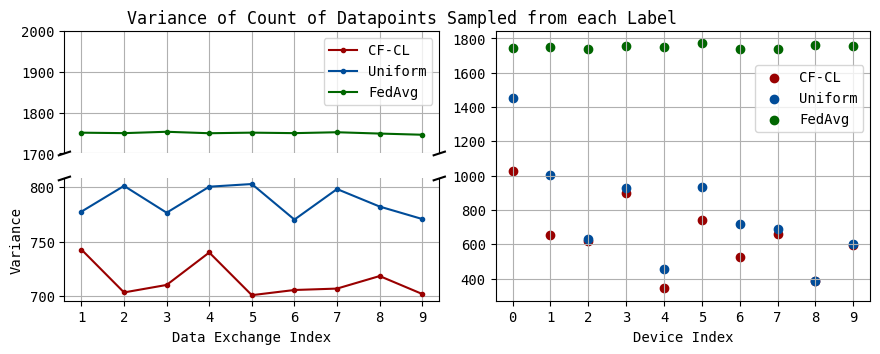}
		\caption{}\label{fig:3c}
	\end{subfigure}
	\begin{subfigure}[t]{0.18\textwidth}
		\centering
		\includegraphics[width=\linewidth,height=3cm]{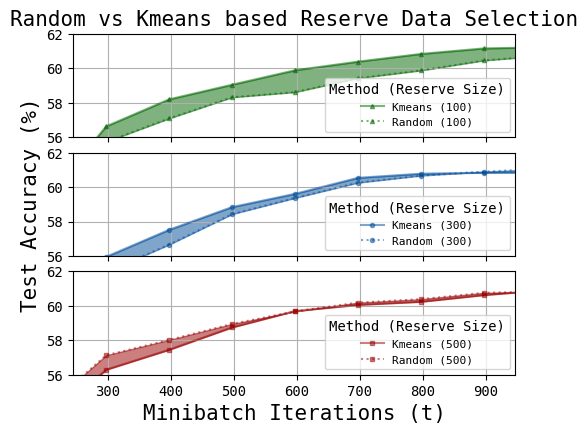}
		\caption{}\label{fig:3d}
	\end{subfigure}
	\begin{subfigure}[t]{0.18\textwidth}
		\centering
		\includegraphics[width=\linewidth,height=3cm]{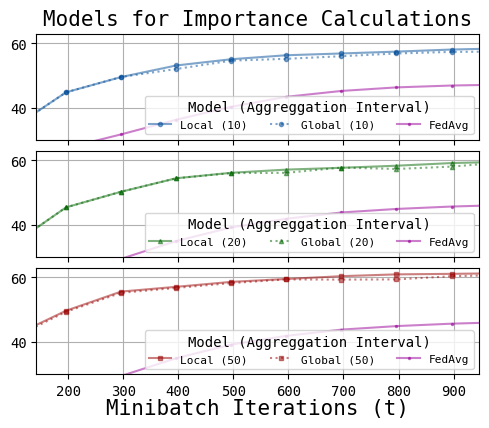}
		\caption{}\label{fig:3e}
	\end{subfigure}
	\begin{subfigure}[t]{0.18\textwidth}
		\centering
		\includegraphics[width=\linewidth,height=3cm]{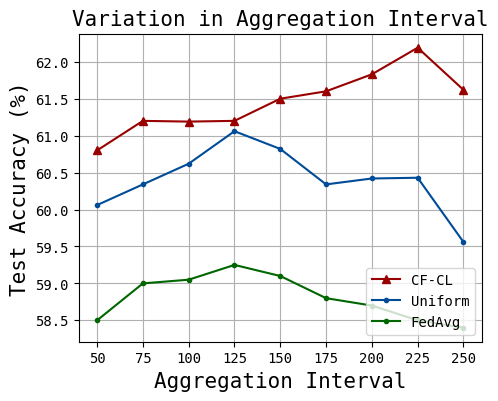}
		\caption{}\label{fig:3f}
	\end{subfigure}	
	\begin{subfigure}[t]{0.36\textwidth}
		\centering
		\includegraphics[width=\linewidth,height=3cm]{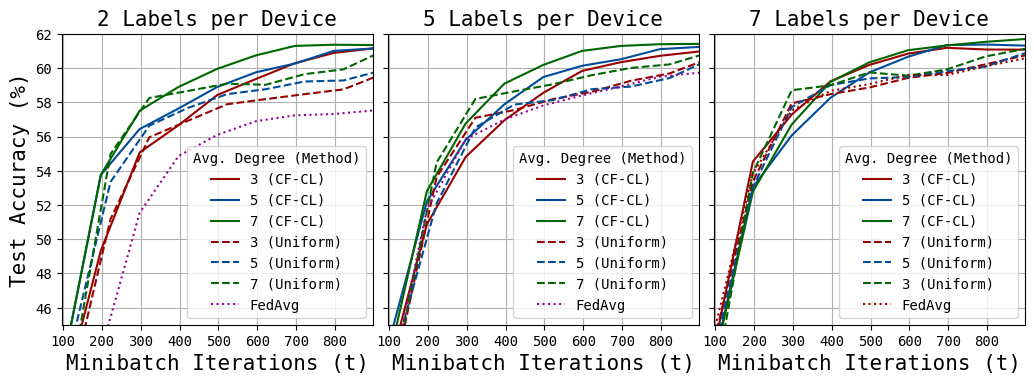}
		\caption{}\label{fig:3g}
	\end{subfigure}
	\vspace{-1.5mm}
	\caption{Simulation results. (a) Latent space visualization with (w) and without (w/) data exchange; (b) Training accuracy of {\tt{CF-CL}} against baselines and delay savings; (c) Variance of sampled data in local SGD iterations across data exchange intervals and devices; (d) Effects of using K-means and random sampling for reserve datapoints selection; (e) Performance of {\tt{CF-CL}} against baselines in large local SGD iteration regime; (f) Effect of connectivity of devices under varying conditions of non-i.i.d. data.}\label{fig:3}
	\vspace{-7mm}
\end{figure*}

\vspace{-1.5mm}
\section{Numerical Experiments}
\label{experiments}


\noindent \textbf{Simulation Setup:} We use Fashion MNIST dataset for our experiments \cite{fmnist}, consisting of $60$K images with $10$ classes. 
We consider a network of {\small$|\mathcal{C}|=10$} devices. We  emulate non-i.i.d. data across devices, where each device has $6$K datapoints from two of $10$ classes. We use a 2-layer convolutional neural networks (CNN), with the first layer having $5$ kernels and the second layer $8$ kernels, each of size $3 \times 3$, followed by a 2 linear layers of sizes $128$ and $64$. The Adam optimizer is used with an initial learning rate of $10^{-4}$ and models are trained for $T = 2500$ local SGD iterations. Data augmentation consists of random resized crops, random horizontal flips, and Gaussian blurs. Unless otherwise stated, we set {\small$K^{\mathsf{Reserve}}_{i \rightarrow j}=500$}, and {\small$K^{\mathsf{Approx}}_j=1000$}, and local K-means with 4 clusters {\small$|\mathcal{L}^{t}_{i\leftarrow j}|=4$}, {\small$\forall i,j$}. Selection temperature is chosen such that it increases linearly as {\small$\lambda^t=6(t/T) + 4$}. We conduct simulations on a desktop with 48GB Tesla-P100 GPU with 128GB RAM.



To obtain the accuracy of predictions, we adopt the linear evaluation \cite{pmlr-v119-chen20j}, and use {\small$\bm{\phi}^t_G$}, {\small$\forall t$}, to train a linear layer {\small$\bm{\theta}$} in a supervised manner on top of {\small$\bm{\phi}^t_G$} to perform a classification at the server.  The linear layer is trained via $1000$ SGD iterations.
As mentioned in Sec.~\ref{intro}, smart data transfer has not been studied in the context of unsupervised federated learning, and literature \cite{wang2021device,zhao2018federated,furl,hosseinalipour2022parallel} has only considered uniform data transfer across the network. Thus, we compare the performance of {\tt CF-CL} against \textit{uniform sampling} (i.e., data points transferred are sampled uniformly at random from the local datasets). We also include the results of classic federated learning (FedAvg), which does not conduct any data transfer across devices.

The communication graph $\mathcal{G}$ is assumed to be random geometric graph (RGG), which is a common model used for wireless peer-to-peer networks. We follow the same procedure as in~\cite{9705093} to create RGG with average node degree $3$. We let devices conduct  {\small$T_a=50$} local SGD iterations and perform data exchange after {\small$T_p=10$} iterations unless otherwise stated.




    \textit{\textbf{Embedding Alignment:}} In Fig.~\ref{fig:3a}, we show the embeddings generated by {\tt CE-CL} (right subplot) and conventional FL (left subplot) at aggregation {\small$\gamma = 50$}. The labels of datapoints are used for color coding. Smart data transfer  in {\tt CE-CL} leads to an embedding space with more separated embeddings, i.e., datapoints with same label are closed to one another. 
    
    
    \textit{\textbf{Speed of Convergence:}}
    In Fig. \ref{fig:3b}, we study the convergence speed of {\tt CF-CL} and baseline methods.
    {\tt{CF-CL}} outperforms all the baselines in terms of convergence speed due to its importance-based data transfers. For example,  {\tt CF-CL} reaches the accuracy of $60\%$ through {$620$} SGD iterations, while uniform takes {$1050$} iterations (i.e., {\tt CF-CL} is $40.95\%$ faster). 
    To further reveal the impact of faster convergence of {\tt CF-CL} on network resource savings, we focus on the latency of model training as a performance metric.
We assume
that transmission rate in D2D and uplink are $1$Mbits/sec with $32$ bits quantization applied on the model parameter and $8$ on datapoints, which results in $45433 \times 32 \div 10^6\approx 1.45$s uplink transmission delay per model parameter exchange ($45433$ is the number of model parameters) and $28\times 28 \times 8 \div  10^6 \approx 6.2$ms D2D delay per data point exchange (each data point is a $28\times 28$ gray-scale image with each pixel taking $256$ values). We also compute the extra computation time of {\tt CF-CL} (i.e., the K-means and importance calculations) and that of uniform sampling and incorporate that into delay computations. The right plot in Fig.~\ref{fig:3b} reveals significant delay savings that {\tt CF-CL} obtains\footnote{FedAvg is omitted from the plot due to its significantly lower performance.} upon reaching various accuracies ($\approx 18.7\%$ on average).


   
   \textit{\textbf{Improving Local Data Homogeneity:}} 
   We studied the variance of count of datapoints sampled from each labels across devices to show the effectiveness of each data sampling/transfer method.
   A more effective data transfer method should ideally result in a more balanced set of datapoints in each local training set. 
   The left subplot of Fig ~\ref{fig:3c} shows the variance with respect to data exchange instance ($\tau$), while the right subplot
   depicts the variance of training datapoints across devices (averaged over training time $T$) for {\tt{CF-CL}}, uniform sampling, and FedAvg. 
   While being fully unsupervised, {\tt{CF-CL}} leads to more homogeneous local training sets across devices (observed through a lower variance), which reveals the practicality of our two-stage probabilistic importance sampling procedure.

    
     \textit{\textbf{Reserve Data Selection:}} 
    In Fig.~\ref{fig:3d}, we investigate the effect of using random sampling of $K^{\mathsf{Reserve}}_{i \rightarrow j}$, $\forall i,j$, data points as reserved datapoints vs. K-means based selection (Sec.~\ref{sec:PushKmeans}), in which device $i$ selects reserve datapoints by running a K-means algorithm on local data with {\small$K^{\mathsf{Reserve}}_{i \rightarrow j}$} clusters, under varying {\small$K^{\mathsf{Reserve}}_{i \rightarrow j}$}.
    From Fig.~\ref{fig:3d},
    performance of {\tt{CF-CL}} improves with selection of reserve data using K-Means. This is because K-means selects datapoints that best approximate the local data distribution. The effect of which is more prominent in extreme cases, e.g., {\small$K^{\mathsf{Reserve}}_{i \rightarrow j}=100$}, {\small$\forall i,j$}, and diminishes as the allowable number of pushed data increases.

    \textit{\textbf{Local vs. Global Models for Importance Calculation:}} In {\tt{CF-CL}}, we chose to use the latest global model {\small$\bm{\phi}_G^{\gamma T_a}$} to conduct data transfer at {\small$\tau T_p$}, where {\small$\gamma T_a \leq \tau T_p < (\gamma+1) T_a$}. 
    An ideal substitute to using the latest global model is to transfer latest/instantaneous local models, which incurs a higher transmission overhead. At each instance of data transfer, {\small$\{\bm{\phi}_i^{\gamma T_p} \}_{i\in\mathcal{C}}$} will be used to calculate importance of data based on the receivers' latest local model in~\eqref{eq:exp_loss}. Fig.~\ref{fig:3e} reveals that, {\tt{CF-CL}} consistently stays on par with this substitute while having significantly lower transmission overhead.
    

    
    
    

    
    \textit{\textbf{Various Aggregation Intervals:}} In Fig.~\ref{fig:3f}, we study model performance in a high local SGD iteration regime ($T_a\in[50,250]$), which results in biased local models. The local model bias is severe for the uniform sampling method, significantly reducing its performance. Comparing the uniform sampling with {\tt CF-CL}, both methods produce `knee' shape plots, i.e., performance improves until a certain point but it drops afterwards due to the local model bias; however, our method can tolerate significantly longer periods of local training. Comparing the knee of the red and blue curves occurring at $225$ and $125$ implies a $~45\%$ less frequent global aggregations, while achieving a better model performance for {\tt CF-CL}. This is particularly useful when we have limitations in uplink transmissions from the devices to the server (e.g., high energy consumption, scarce uplink bandwidth), where low-power and short-rage D2D data transfers can be utilized.
    
    
    \textit{\textbf{Local Data Availability:}} Fig.~\ref{fig:3g} shows different scenarios of non-i.i.d data with varying connectivity between devices. We vary the number of labels in each device's local dataset and show that as the local data distributions become more non-i.i.d. (i.e., fewer labels), the speed of convergence of the methods drops due to more biased local models. 
    In such cases, higher connectivity significantly improves the performance, as a higher connectivity allows for exposure of local datasets to a more diverse set of datapoints resulting in less biased local models. Also,
    our method consistently exhibits the best performance across different non-i.i.d. settings with the largest gap to baselines upon having extremely non-i.i.d. data across the devices (i.e., $2$ labels), which addresses one of the biggest challenges of FL across wireless edge devices~\cite{wang2019adaptive}.
    

    

\vspace{-1.5mm}
\section{Conclusion}
\vspace{-.2mm}
\noindent We proposed Cooperative Federated unsupervised Contrastive Learning ({\tt CF-CL}). In {\tt CF-CL} devices learn representations of unlabeled data and engage in cooperative smart data push-pull to eliminate the local model bias. We proposed a randomized data importance estimation and subsequently developed a two-staged probabilistic data sampling scheme across the devices. Through numerical simulations, we studied the model training behavior of {\tt CF-CL} and showed that it outperforms the baseline methods in terms of accuracy and efficiency.

\vspace{-2mm}
\bibliographystyle{IEEEtran}
\bibliography{fcl-refs.bib} 

\end{document}